\documentclass[twoside,11pt]{article}

\usepackage{jmlr2e}

\usepackage[utf8]{inputenc} 
\usepackage[T1]{fontenc}    
\usepackage{hyperref}       
\usepackage{url}            
\usepackage{booktabs}       
\usepackage{amsfonts}       
\usepackage{nicefrac}       
\usepackage{amsmath,amsfonts,amssymb,algorithm,algorithmic}
\usepackage{microtype}      
\usepackage{lipsum}		
\usepackage{graphicx}
\usepackage{amsmath}
\usepackage{natbib}
\usepackage{caption}
\usepackage{subcaption}
\usepackage{doi}
\usepackage{tikz}
\usetikzlibrary{positioning}
\usepackage{amsmath}
\DeclareMathOperator*{\argmax}{arg\,max}
\jmlrheading{1}{}{}{}{}{}

\firstpageno{1}

\begin{document}

\title{Expert-guided Bayesian Optimisation for Human-in-the-loop Experimental Design of Known Systems}

\author{\name Tom Savage \email t.savage@imperial.ac.uk \\
       \addr Imperial College London; The Alan Turing Institute
       \AND
       \name Ehecatl Antonio del Rio Chanona \email a.del-rio-chanona@imperial.ac.uk \\
       \addr Imperial College London}


\maketitle

\begin{abstract}

Domain experts often possess valuable physical insights that are overlooked in fully automated decision-making processes such as Bayesian optimisation.
In this article we apply high-throughput (batch) Bayesian optimisation alongside anthropological decision theory to enable domain experts to influence the selection of optimal experiments. 
Our methodology exploits the hypothesis that humans are better at making discrete choices than continuous ones and enables experts to influence critical early decisions. 
At each iteration we solve an augmented multi-objective optimisation problem across a number of alternate solutions, maximising both the sum of their utility function values and the determinant of their covariance matrix, equivalent to their total variability. 
By taking the solution at the knee point of the Pareto front, we return a set of alternate solutions at each iteration that have both high utility values and are reasonably distinct, from which the expert selects one for evaluation.
We demonstrate that even in the case of an uninformed practitioner, our algorithm recovers the regret of standard Bayesian optimisation. 

\end{abstract}

\section{Introduction}

\begin{figure}[htb!]
    \centering
    \includegraphics[width=0.9\textwidth]{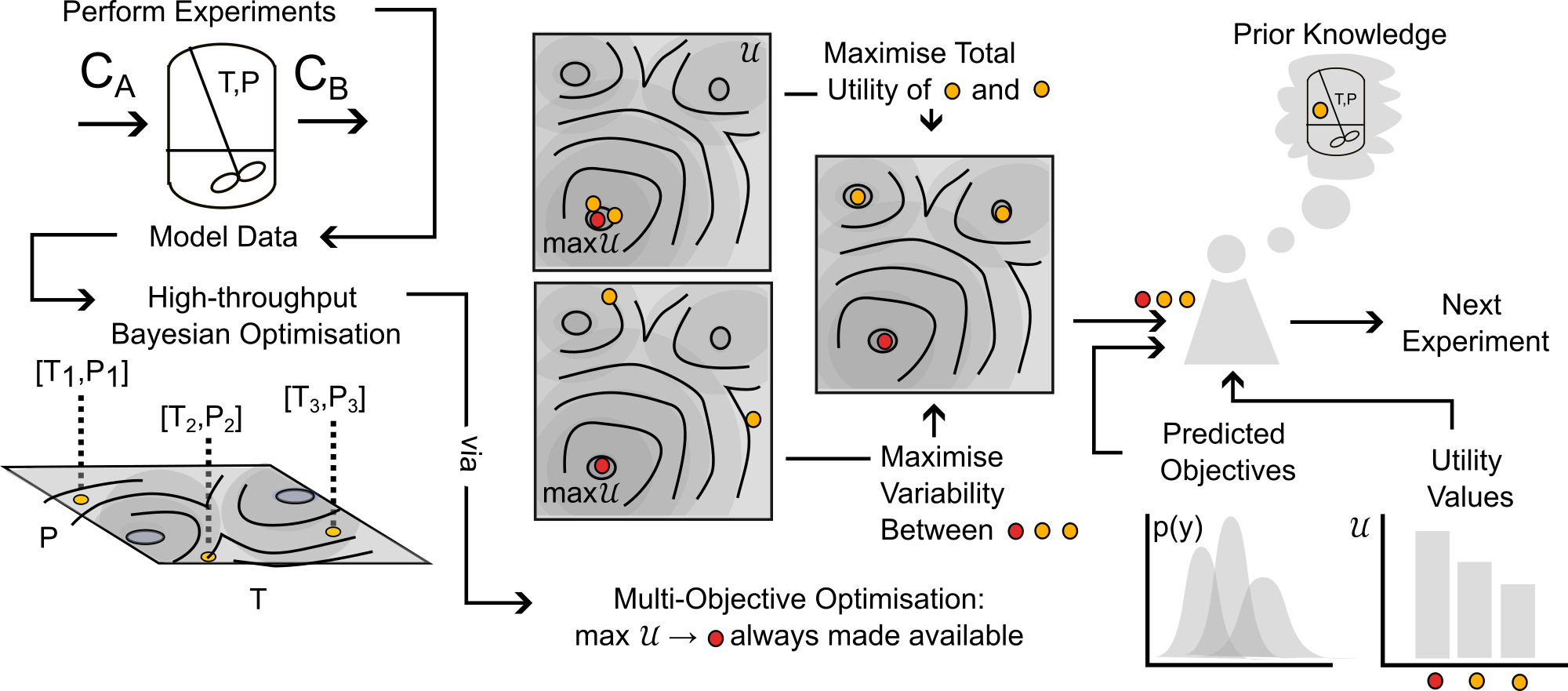}
    \caption{Overview of our methodology, where an augmented batch Bayesian optimisation problem is solved using multi-objective optimisation, providing an expert with a set of alternate solutions.}
    \label{fig:overview}
\end{figure}

Bayesian optimisation has been successfully applied in a number of complex domains including engineering systems where derivatives are often not available, such as those that involve simulation or propriety software. 
By removing the human from decision-making processes in favour of maximising statistical quantities such as expected improvement, complex functions can be optimised in an efficient number of samples. 
However, these engineering systems are often engaged with by domain experts such as engineers or chemists, and as such the behaviour of the underlying function cannot be considered completely unknown a-priori. 
Therefore, there exists significant scope to take advantage of the benefits of Bayesian optimisation in optimising expensive derivative-free problems, whilst enabling domain experts to inform the decision-making process, putting the human back into the loop. 
By providing an optimal set of alternatives to an expert to select their desired evaluation, we ensure that any one choice presents information gain about the optimal solution.
Simultaneously, we ensure the choices are distinct enough to avoid the human making an effective gradient calculation. 
Alternative solution information such as utility function value, predictive output distribution and visualisations are provided to the expert as a pseudo-likelihood. 
The decision-maker then effectively performs discrete Bayesian reasoning, internally conditioning the provided information with their own prior expertise and knowledge of the solutions provided. 
In addition to improved convergence (depending on the ability of the domain expert), our methodology enables improved interpretability in higher dimensions, as the decision-maker has the final say in what is evaluated. 
Our approach works with any utility function and NSGA-II \citep{Deb2002} is applied for multi-objective optimisation, efficiently handling the non-convex utility-space.  
Figure \ref{fig:overview} demonstrates our methodology\footnote{Code can be located at \url{https://github.com/trsav/HITL-BO}.}.

By allowing an expert to influence the experimental design through a discrete decision step, we mitigate the expert needing to make continuous decisions throughout, and do not rely on an expert `prior' of the global optimum that necessarily must be defined before optimisation. 
Approaches that rely on an expert-defined prior may need to redefine this at significant human-cost throughout optimisation in light of new information.
Similarly, the expert has no influence over the actual solutions evaluated and the optimisation is merely weighted towards broad regions in solution-space.

\section{Previous Work}

\citet{Kanarik2023} demonstrated that experts improve the initial convergence in Bayesian optimisation for semiconductor processes. 
However, this can be counterproductive in later stages. 
Algorithmic integration of expert knowledge has been explored \citep{Liu2022,hvarfner2022pibo}. 
\citet{hvarfner2022pibo} and \citet{Ramachandran2020} use user-defined priors to weight the acquisition function, while \citet{Liu2022} diminishes this weight over time. 
These methods are static and don't allow real-time expert input. 
\citet{BOMuse} allows continuous expert involvement, using a linear Gaussian process to approximate human intuition, achieving sub-linear regret bounds. 
\citet{av2022human} and \citet{robot} present similar frameworks, offering alternative solutions for evaluation, with the expert making the final decision latterly in a molecular design setting.

\section{Method}

We first maximise a given utility function $\mathcal{U}$ for a given dataset $\mathcal{D}_t:= \{(\mathbf{x}_i,y_i)\}_{i=1}^t$:
\begin{align}\label{standard_bo}
   \mathbf{x}^* = \argmax_{x\in\mathcal{X}\subseteq\mathbb{R}^n} \; \mathcal{U}(x),
\end{align}
resulting in the optimal next evaluation, $\mathbf{x}^*$, in a utility sense.
Let $p$ be the number of alternate solutions provided to the expert and construct the decision variable matrix $\mathbf{X} \in \mathbb{R}^{(p-1)\times n}$ by concatenating $p-1$ alternate solutions $\mathbf{X} := [\mathbf{x}_1,\dots,\mathbf{x}_{p-1}]$.
We then define the high-throughput (batch) utility function $\hat{\mathcal{U}}$ which is specified as the sum of the individual utilities of alternate solutions within $\mathbf{X}$
\begin{align}
    \hat{\mathcal{U}}(\mathbf{X}) = \sum_{i=0}^{p-1} \mathcal{U}(\mathbf{X}_i).
\end{align}
Similarly, we introduce $\hat{\mathcal{S}}$ as a measure for capturing the variability among both the optimal and alternative solutions.
Specifically, let $\hat{\mathcal{S}}$ be the determinant of the covariance matrix $K_{\mathbf{X}_{\text{aug}}}$ for the augmented set $ \mathbf{X}_{\text{aug}}= \mathbf{X} \cup \mathbf{x}^*$:
\begin{align*}
 \hat{\mathcal{S}}(\mathbf{X},\mathbf{x}^*) &= |K_{\mathbf{X_{\text{aug}}}}| \\
 K_{\mathbf{X}_{\text{aug}}} &= [k(\mathbf{X}_{\text{aug},i},\mathbf{X}_{\text{aug},j})]^p_{i,j=1}
\end{align*}
$\hat{\mathcal{S}}$ quantifies the `volume of information' spanned by the alternative solutions $\mathbf{X}$ as well as the optimal solution $\mathbf{x}^*$.
Maximising $\hat{\mathcal{U}}$ will result in all alternative solutions proposed being the same as $\mathbf{x}^*$, that is $[\mathbf{x}^*_1,\dots,\mathbf{x}^*_{p-1}]$.
Contrary to this, maximising $\hat{\mathcal{S}}$ will result in a set of solutions that are maximally-spaced both with respect to other alternatives, but also $\mathbf{x}^*$.
At iteration $t$, we then solve the following multi-objective optimisation problem:
\begin{align}\label{multi-objective}
    [\mathbf{X}^*_1,\dots,\mathbf{X}^*_m] = \max_{\mathbf{X}} \; \left(\hat{\mathcal{U}}(\mathbf{X};\mathcal{D}_t),\hat{\mathcal{S}}(\mathbf{X},\mathbf{x}^*)\right),
\end{align}
resulting in a set of $m$ solutions along the Pareto front of both objectives. 
From this we define $\mathbf{X}^*_{k}$ as the solution at knee-point of the Pareto front. 
The $p-1$ solutions contained within $\mathbf{X}^*_k$ optimally trade off the sum of their utility values, with their variability. 
This ensures that when provided to an expert, alongside $\mathbf{x}^*$, any individual solution will have high expected information gain, and the solutions themselves will be distinct enough to ensure the expert isn't made to make an effective gradient calculation (i.e. solutions are very close and practically indistinguishable).   
The practitioner is then made to choose a solution to evaluate from this set of alternatives. 
To do so, they are provided with information such as the utility value of each solution, expected output distributions (obtained from the Gaussian process), and information regarding previous solutions that they may wish to draw upon. 
In doing so, the practitioner effectively performs an internal discrete Bayesian reasoning, conditioning previous prior information and expert opinion with the mathematical quantities provided to make an informed decision.
Our algorithm can be located within the Appendix.

\begin{figure}[htb!]
    \centering
        \begin{subfigure}[b]{\textwidth}
         \centering
        \includegraphics[width=\textwidth]{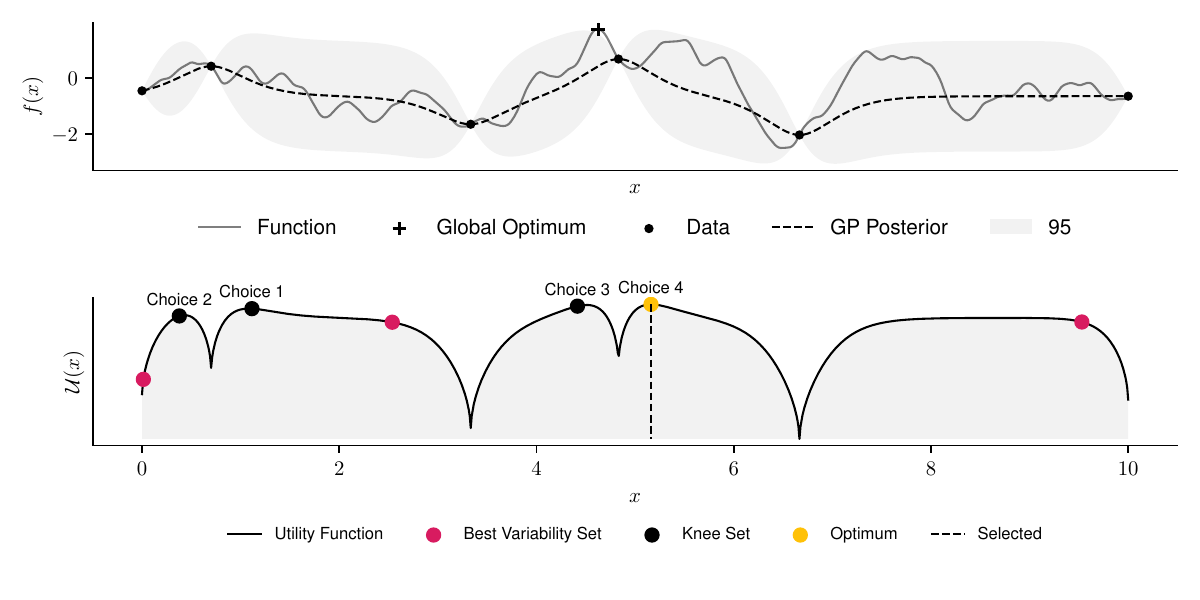}
         \caption{The objective function and utility function after 6 function evaluations. The 3 alternative solutions that maximise the solution distance can be seen in red, whilst the black solutions denote those contained within the knee-solution of the high-throughput multi-objective problem. The yellow optimal solution is included alongside these two alternatives to an expert. In this case choice 4 is selected randomly from the 3 alternatives and the optimum.}
         \label{b_1_aq}
     \end{subfigure}
        \begin{subfigure}[b]{\textwidth}
         \centering
        \includegraphics[width=\textwidth]{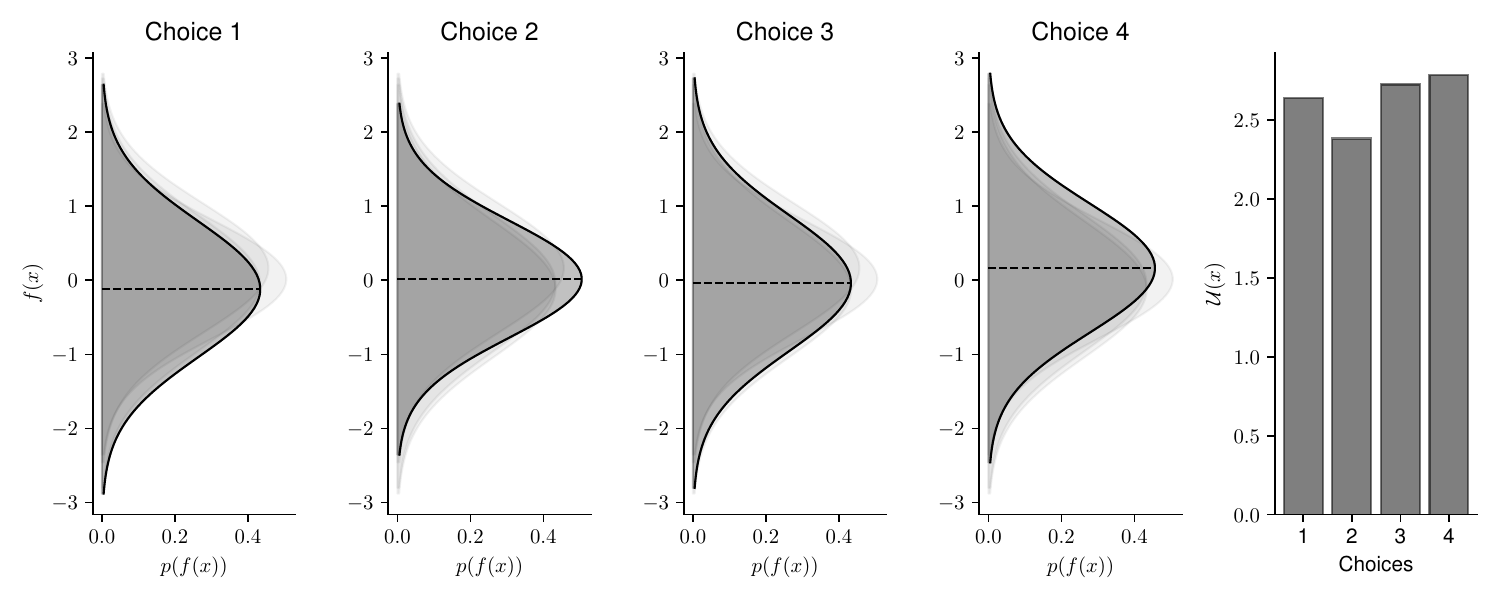}
         \caption{The information provided to the expert regarding the three alternative solutions. In this case choice 1 and choice 3 have relatively similar utility values and predicted output distributions to choice 4 (the optimal of the acquisition function). The expert is then allowed to distinguish between these similar solutions in a way the computer cannot through their prior domain knowledge. By conditioning their prior information on the given values, the expert is effectively performing internal Bayesian reasoning.}
         \label{b_1_choices}
     \end{subfigure}
     \caption{A standard iteration of our approach on a one-dimensional case-study.}
    \label{behaviour_1}
\end{figure}

Figure \ref{behaviour_1} demonstrates the intended behaviour of our approach. We present a one-dimensional case study, optimising a function obtained through sampling a Gaussian process prior, specified by a Mat\'ern 5/2 kernel with lengthscale $l=0.5$. In this case study we provide 3 alternatives to an expert, who's choice we select randomly.
We provide details of optimisation methods and hyper-parameters within the Appendix.

\section{Computational Results \& Discussion}

To assess our approach, we benchmark it against standard Bayesian optimisation.
In order to incorporate and assess human interaction in an automated manner, we hypothesise a number of different human behaviours.
The `Expert' practitioner represents an ideal, where the best solution (that is the one with the highest true objective value) is always selected. 
Equivalently, to test the performance of our approach under the influence of a practitioner with misaligned knowledge, we present an `Adversarial' practitioner who consistently selects the solution with the lowest true objective value.
In addition, we present a probabilistic practitioner, who selects the solution with the best true objective value with some probability. 
Finally, we present the behaviour of a `Trusting' practitioner who selects the solution with the largest utility (as these values are presented), equivalent to standard Bayesian optimisation as this solution is obtained through standard single objective optimisation (Eq. \ref{standard_bo}). 
In practice, the expert will condition the information provided with their prior beliefs. 
In our approach this includes information regarding the expected distribution of the objective of each solution, as well as the utility value of each solution. 
Whilst we cannot benchmark real human behaviour due to the random nature of the objective functions, and practical issues, the behaviours described summarise key aspects in order to generate useful insights into our approach, we leave this for future work.
The human behaviours applied are summarised within Table \ref{behav}.

\begin{table}[h]
\centering
\caption{Human Behaviours Applied for Benchmarking}
\label{behav}
\renewcommand{\arraystretch}{1.5} 
\begin{tabular}{ll}
\hline
\textbf{Behaviour Type} & \textbf{Description}                                                                                                                                       \\ \hline
Expert                                  & Selects the solution with the best true function value.                                                                                                                     \\
Adversarial                             & Selects the solution with the worst true function value.                                                                                                                    \\
Trusting                                & Selects the solution with the maximum utility value.                                                                                                                        \\
\(p(\text{Best})\)                      & \begin{tabular}[c]{@{}l@{}}Selects the solution with the best true function value with \\ probability \(p(\text{Best})\), otherwise selects a random solution.\end{tabular} \\ \hline
\end{tabular}
\end{table}

We perform optimisation over 50 functions, each representing a sample from a Gaussian process prior with lengthscale 0.04 within unit bounds using the upper-confidence bound (UCB) utility function.
Figure \ref{toy_problem} demonstrates the average and standard deviation of simple regret, and average regret (both defined within \cite{Garnett2023}) for each human behaviour across 1D and 2D objective functions.  
Results for 5D, and specific functions can be located within the Appendix as well as examples of sampled objective functions. 

\begin{figure}[htb!]
    \centering
    \begin{subfigure}[b]{\textwidth}
    \centering
    \includegraphics[width=0.9\textwidth]{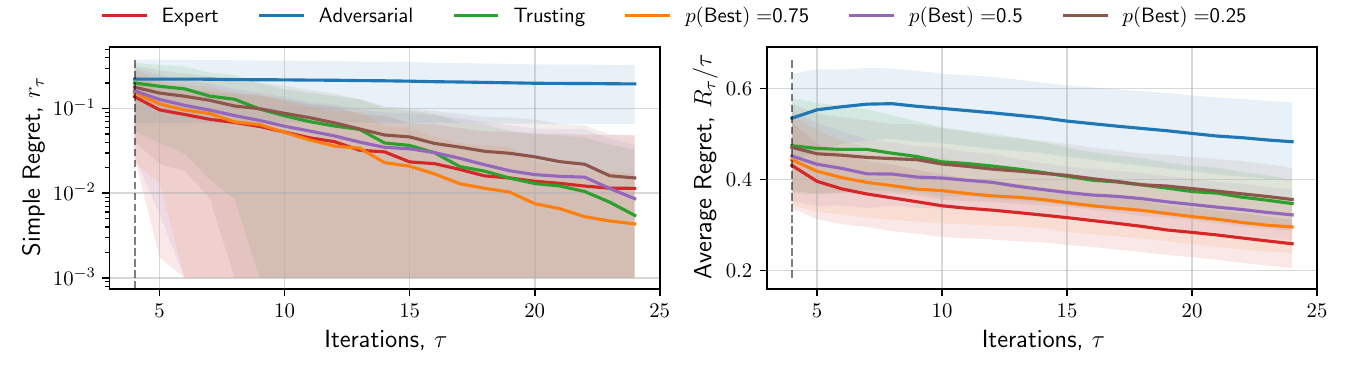}
    \caption{Average regret quantities over 1D objective functions.}
    \label{1D}
    \end{subfigure}
    \begin{subfigure}[b]{\textwidth}
    \centering
    \includegraphics[width=0.9\textwidth]{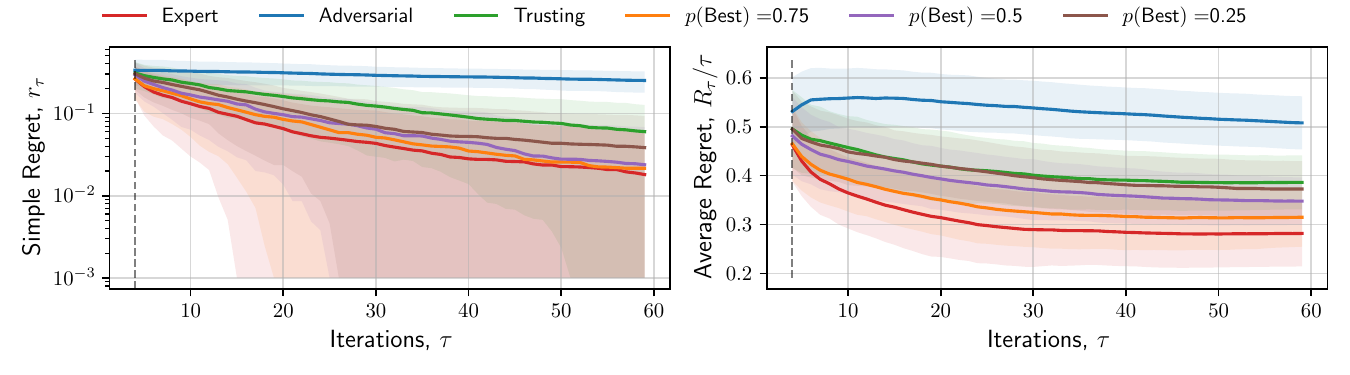}
    \caption{Average regret quantities over 2D objective functions.}
    \label{2D}
    \end{subfigure}
    \caption{Regret expectation over 50 functions, $f \sim \mathcal{GP}(\mu \equiv 0, K_M (d,\nu = 0.04))$ where $K_M$ is the Mat\'ern 5/2 kernel function. 4 alternate choices are presented to the practitioner, and the utility function $\mathcal{U}(x)$ used is the upper-confidence bound.}
    \label{toy_problem}
\end{figure}

The expectation of average regret tends towards zero for all behaviours, indicating empirical convergence. 
Focusing on results from the set of 1D functions, an `Expert' provides improved convergence than standard Bayesian optimisation (`Trusting') throughout all iterations, with benefits diminishing throughout the later stages, where the standard automated approach recovers the `Expert' average simple regret, confirming previous observations regarding the importance of human input throughout earlier iterations, and conversely diminishing importance of expert opinion during `fine-tuning' \citep{Kanarik2023}.
Improved convergence of simple regret occurs in cases where the practitioner selects the `best' solution from a set 75\%, 50\%, and to a lesser extent 25\% of the time out of 4 available choices, similarly reflected in trends across average regret.
The results demonstrated in Figure \ref{toy_problem} indicate the potential for our approach to improve the convergence of Bayesian optimisation even in cases where the practitioner is correct about a decision only partially.
When the expert makes a random selection ($p(\text{Best}) = 0.25$, for 4 alternate solutions), standard Bayesian optimisation convergence is recovered, indicating the effectiveness of the choices presented by asking a practitioner to select between distinct solutions, each of which individually has a high utility value.
This is reflected throughout 1D, 2D and to a lesser extent 5D functions.
Only in the case where the practitioner actively selects the worst solution (i.e. they are misaligned), is performance worse. 
In higher-dimensions an adversarial practitioner performs worse, as they are performing inefficient space-filling in an increasingly larger volume before good solutions are found.

The methodology we present may also be interpreted as an approach for high-throughput/batch Bayesian optimisation, with an additional preference for solutions that are well-distributed throughout the search space.
Our intention for future work is to benchmark it against other existing batch Bayesian optimisation methodologies \citep{local_pen}, including those with similar multi-objective formulations \citep{Bischl2014,Habib2016,robot}. 
We will also investigate to what extent large-language models can perform the selection step.

\newpage

\vskip 0.2in
\bibliography{sample}

\begin{thebibliography}{13}
\providecommand{\natexlab}[1]{#1}
\providecommand{\url}[1]{\texttt{#1}}
\expandafter\ifx\csname urlstyle\endcsname\relax
  \providecommand{\doi}[1]{doi: #1}\else
  \providecommand{\doi}{doi: \begingroup \urlstyle{rm}\Url}\fi

\bibitem[Bischl et~al.(2014)Bischl, Wessing, Bauer, Friedrichs, and Weihs]{Bischl2014}
Bernd Bischl, Simon Wessing, Nadja Bauer, Klaus Friedrichs, and Claus Weihs.
\newblock {{MOI}-{MBO}: Multiobjective Infill for Parallel Model-Based Optimization}.
\newblock In \emph{Lecture Notes in Computer Science}, pages 173--186. Springer International Publishing, 2014.
\newblock \doi{10.1007/978-3-319-09584-4_17}.

\bibitem[Deb et~al.(2002)Deb, Pratap, Agarwal, and Meyarivan]{Deb2002}
K.~Deb, A.~Pratap, S.~Agarwal, and T.~Meyarivan.
\newblock {A fast and elitist multiobjective genetic algorithm: {NSGA}-{II}}.
\newblock \emph{{IEEE} Transactions on Evolutionary Computation}, 6\penalty0 (2):\penalty0 182--197, April 2002.
\newblock \doi{10.1109/4235.996017}.

\bibitem[Garnett(2023)]{Garnett2023}
Roman Garnett.
\newblock \emph{{Bayesian Optimization}}.
\newblock Cambridge University Press, January 2023.
\newblock \doi{10.1017/9781108348973}.

\bibitem[González et~al.(2015)González, Dai, Hennig, and Lawrence]{local_pen}
Javier González, Zhenwen Dai, Philipp Hennig, and Neil~D. Lawrence.
\newblock {Batch Bayesian Optimization via Local Penalization}.
\newblock 2015.
\newblock \doi{10.48550/ARXIV.1505.08052}.

\bibitem[Gupta et~al.(2023)Gupta, Shilton, A, Ryan, Abdolshah, Le, Rana, Berk, Rashid, and Venkatesh]{BOMuse}
Sunil Gupta, Alistair Shilton, Arun~Kumar A, Shannon Ryan, Majid Abdolshah, Hung Le, Santu Rana, Julian Berk, Mahad Rashid, and Svetha Venkatesh.
\newblock {BO-Muse: A human expert and AI teaming framework for accelerated experimental design}.
\newblock 2023.
\newblock \doi{10.48550/ARXIV.2303.01684}.

\bibitem[Habib et~al.(2016)Habib, Singh, and Ray]{Habib2016}
Ahsanul Habib, Hemant~Kumar Singh, and Tapabrata Ray.
\newblock {A multi-objective batch infill strategy for efficient global optimization}.
\newblock In \emph{2016 {IEEE} Congress on Evolutionary Computation ({CEC})}. {IEEE}, July 2016.
\newblock \doi{10.1109/cec.2016.7744341}.

\bibitem[Hvarfner et~al.(2022)Hvarfner, Stoll, Souza, Lindauer, Hutter, and Nardi]{hvarfner2022pibo}
Carl Hvarfner, Danny Stoll, Artur Souza, Marius Lindauer, Frank Hutter, and Luigi Nardi.
\newblock {$\pi$BO: Augmenting Acquisition Functions with User Beliefs for Bayesian Optimization}.
\newblock 2022.

\bibitem[Kanarik et~al.(2023)Kanarik, Osowiecki, Lu, Talukder, Roschewsky, Park, Kamon, Fried, and Gottscho]{Kanarik2023}
Keren~J. Kanarik, Wojciech~T. Osowiecki, Yu~Lu, Dipongkar Talukder, Niklas Roschewsky, Sae~Na Park, Mattan Kamon, David~M. Fried, and Richard~A. Gottscho.
\newblock {Human-machine collaboration for improving semiconductor process development}.
\newblock \emph{Nature}, 616\penalty0 (7958):\penalty0 707--711, March 2023.
\newblock \doi{10.1038/s41586-023-05773-7}.

\bibitem[Kumar et~al.(2022)Kumar, Rana, Shilton, and Venkatesh]{av2022human}
A~V Kumar, Arun, Santu Rana, Alistair Shilton, and Svetha Venkatesh.
\newblock {Human-AI Collaborative Bayesian Optimisation}.
\newblock \emph{Advances in Neural Information Processing Systems}, 35:\penalty0 16233--16245, 2022.

\bibitem[Liu(2022)]{Liu2022}
Peng Liu.
\newblock {Human-in-the-loop Bayesian Optimization with No-Regret Guarantees}.
\newblock October 2022.
\newblock {Available at SSRN: \url{https://ssrn.com/abstract=4262945}}.

\bibitem[Maus et~al.(2022)Maus, Wu, Eriksson, and Gardner]{robot}
Natalie Maus, Kaiwen Wu, David Eriksson, and Jacob Gardner.
\newblock {Discovering Many Diverse Solutions with Bayesian Optimization}.
\newblock 2022.
\newblock \doi{10.48550/ARXIV.2210.10953}.

\bibitem[Ramachandran et~al.(2020)Ramachandran, Gupta, Rana, Li, and Venkatesh]{Ramachandran2020}
Anil Ramachandran, Sunil Gupta, Santu Rana, Cheng Li, and Svetha Venkatesh.
\newblock {Incorporating expert prior in Bayesian optimisation via space warping}.
\newblock \emph{Knowl. Based Syst.}, 195:\penalty0 105663, May 2020.
\newblock \doi{10.1016/j.knosys.2020.105663}.

\bibitem[Zhu et~al.(1997)Zhu, Byrd, Lu, and Nocedal]{zhu1997algorithm}
Ciyou Zhu, Richard~H Byrd, Peihuang Lu, and Jorge Nocedal.
\newblock {Algorithm 778: L-BFGS-B: Fortran subroutines for large-scale bound-constrained optimization}.
\newblock \emph{ACM Transactions on mathematical software (TOMS)}, 23\penalty0 (4):\penalty0 550--560, 1997.

\end{thebibliography}

\appendix

\section*{Appendix A. (Algorithm)}
\label{app:theorem}

Algorithm \ref{alg:Human-Informed_BO} demonstrates our approach. Our approach allows for an expert to include a number of continuous solutions within the initial dataset $\mathcal{D}$, which design of experiments can be performed around.
Additionally the algorithm may be used alongside other approaches which allow for an expert to specify a distinct solution to be evaluated at every iteration.

\begin{algorithm}
\caption{Expert-Guided Bayesian Optimization}
\label{alg:Human-Informed_BO}
\begin{algorithmic}[1]
    \STATE \textbf{Initialize:} Objective function \(f\), Initial data \(\mathcal{D}\), Domain \(\mathcal{X}\) 
    \STATE Alternative choices \(p\), Utility function \(\hat{\mathcal{U}}\), Variability function \(\hat{\mathcal{S}}\), Termination Criteria \\
    \vspace{2mm} 

    \WHILE{Termination Criteria is False}
        \STATE \(\mathbf{x}^* \leftarrow \argmax_{x\in\mathcal{X}} \mathcal{U}(x)\) \texttt{ // Standard Bayesian Optimization Step} \\
        \vspace{2mm} 

        \STATE \texttt{\textbf{// Multi-Objective High-Throughput Optimization}} 
        \STATE \([\mathbf{X}^*_1,\dots,\mathbf{X}^*_m] \leftarrow \max_{\mathbf{X}} (\hat{\mathcal{U}}(\mathbf{X};\mathcal{D}), \hat{\mathcal{S}}(\mathbf{X},\mathbf{x}^*))\) \\
        \vspace{2mm} 

        \STATE \(\mathbf{X}^*_k \leftarrow \text{knee}([\mathbf{X}^*_1,\dots,\mathbf{X}^*_m])\) \texttt{ // Select Knee-solution} \\
        \vspace{2mm} 

        \STATE \texttt{\textbf{// Expert selects from alternatives and standard optimal}} 
        \STATE \(\mathbf{x}_{\text{eval}} \leftarrow \underset{\mathbf{x} \in \mathbf{X}^*_k\cup \mathbf{x}^*}{\text{expert}} \mathbf{x}\) \\
        \vspace{2mm} 

        \STATE \(\mathcal{D} \leftarrow \{(\mathbf{x}_{\text{eval}}, f(\mathbf{x}_{\text{eval}}))\} \cup \mathcal{D}\) \texttt{ // Update dataset} \\
    \ENDWHILE
\end{algorithmic}
\end{algorithm}

\section*{Appendix B. (Optimisation Details)}

Throughout this paper we apply 4 alternative solutions at each iteration (one of which will be the optimum of the utility function). 
Throughout all optimisation problems we generate an initial sample of 4 experiments, distributed via a static Latin hypercube design-of-experiments.
When training the Gaussian process that models the objective-space we perform a multi-start of 8 gradient-based optimisation runs using Adam with a learning rate of 1e-3 for 750 iterations to determine GP hyperparameters. 
To solve the multi-objective augmented optimisation problem, resulting in a Pareto set of sets of alternative solutions we run NSGA-II \citep{Deb2002} for 150 iterations, with a population size of 100, 30 offspring at each iteration, a 0.9 probability of crossover, and 20 mutations per iteration. 
To generate the single optimum of the acquisition function, we perform a multi-start of 36 gradient-based optimisation runs using L-BFGS-B \citep{zhu1997algorithm} with a tolerance of 1e-12 and a maximum iterations of 500. 
For the 1D, 2D, and 5D expectations over functions stemming from samples of a Gaussian process prior (with lengthscale 0.3), the lower and upper bounds used are 0 and 10 respectively for each dimension.
All other bounds for functions can be located at \url{http://www.sfu.ca/~ssurjano/optimization.html}.

\section*{Appendix C. (Example Sampled Objective Functions)}

\begin{figure}[H]
    \centering
    \includegraphics[width=0.85\textwidth]{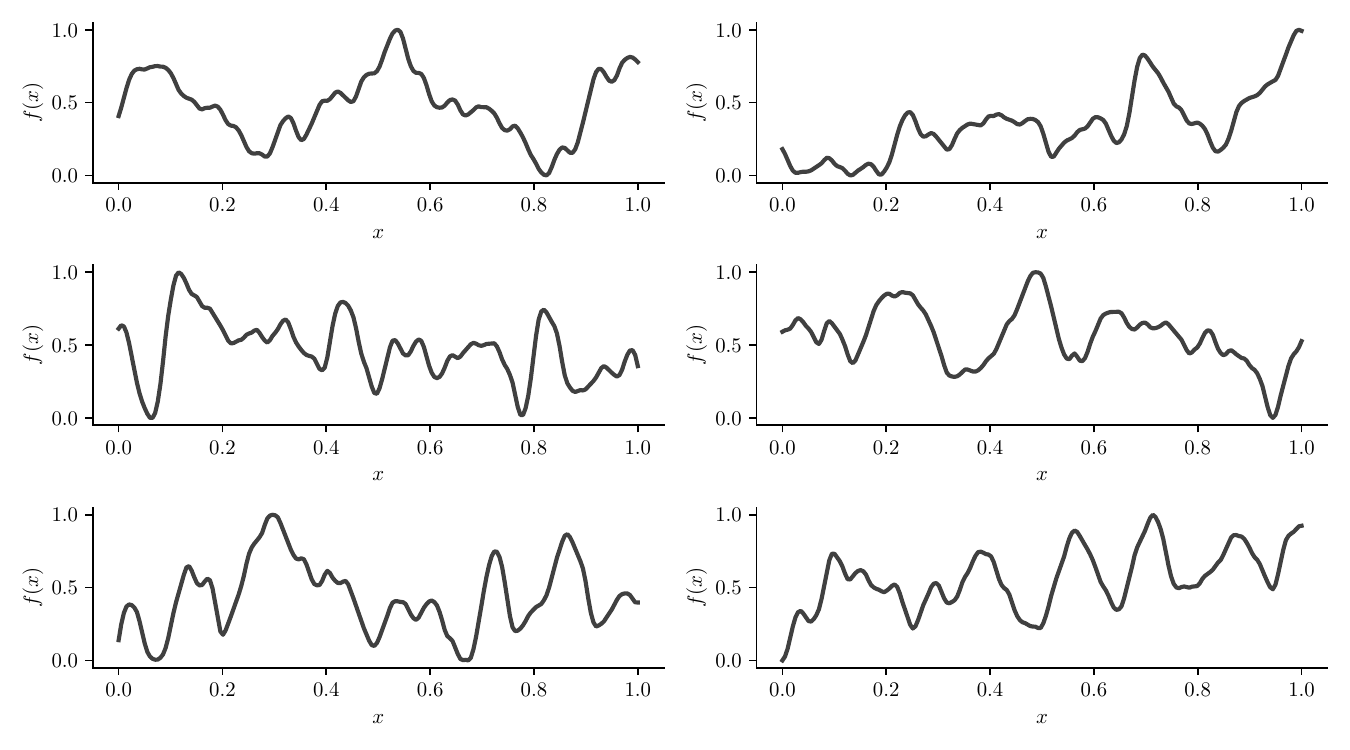}
    \caption{Functions sampled from a Gaussian process prior used for 1D benchmarking $f \sim \mathcal{GP}(\mu \equiv 0, K_M (d,\nu = 0.04))$ where $K_M$ is the Mat\'ern 5/2 kernel function.}
\end{figure}

\begin{figure}[H]
    \centering
    \includegraphics[width=0.85\textwidth]{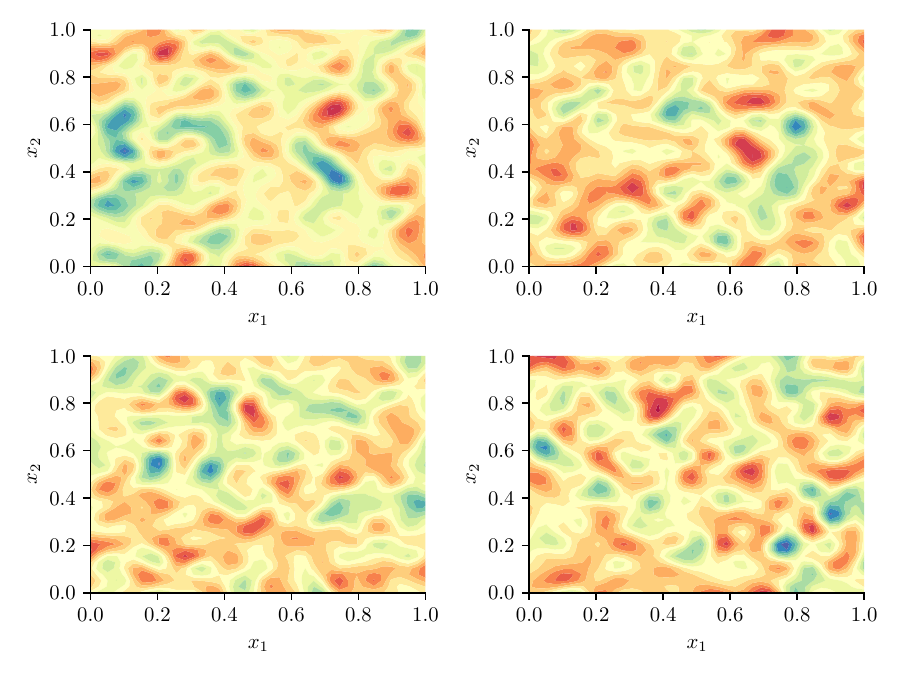}
    \caption{Functions sampled from a Gaussian process prior used for 2D benchmarking $f \sim \mathcal{GP}(\mu \equiv 0, K_M (d,\nu = 0.04))$ where $K_M$ is the Mat\'ern 5/2 kernel function.}
\end{figure}

\section*{Appendix D. (Regret Plots)}

\begin{figure}[H]
    \centering
    \includegraphics[width=\textwidth]{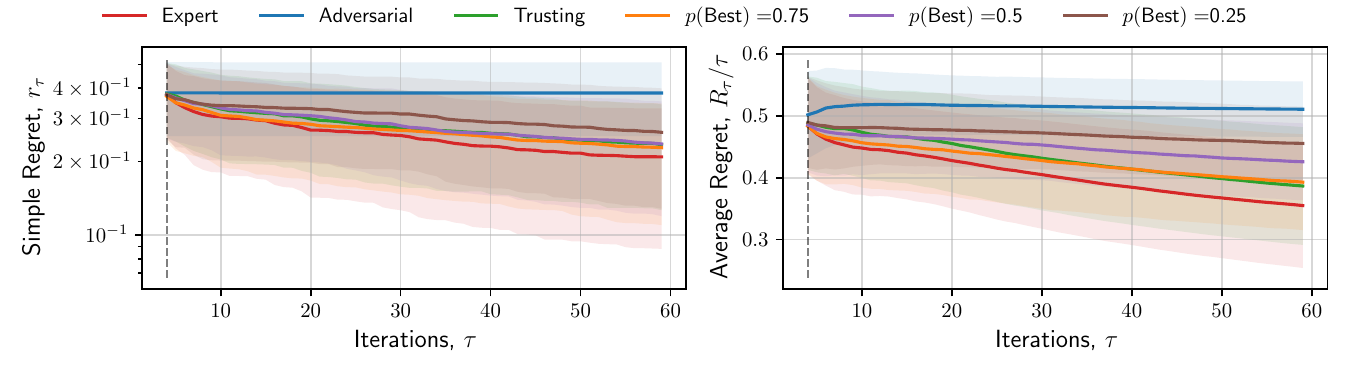}
    \caption{Regret expectation over 50 5D functions ($d=5$), $f \sim \mathcal{GP}(\mu \equiv 0, K_M (d,\nu = 0.3))$ where $K_M$ is the Mat\'ern 5/2 kernel function. 4 alternate choices are presented to the practitioner, and the utility function $\mathcal{U}(x)$ used is the upper-confidence bound.}
    \label{toy_problem}
\end{figure}

We subsequently present the expectation of regret for a number of functions across all human behaviours.
We present 4 alternate choices are presented to the hypothetical practitioner, and the utility function $\mathcal{U}(x)$ used is the upper-confidence bound.
Each function is optimised 16 times, across a number of random initialisations, and average regret quantities are plotted.
Optimisation details are presented in the previous Appendix section. 

\begin{figure}[htb!]
    \centering
    \includegraphics[width=\textwidth]{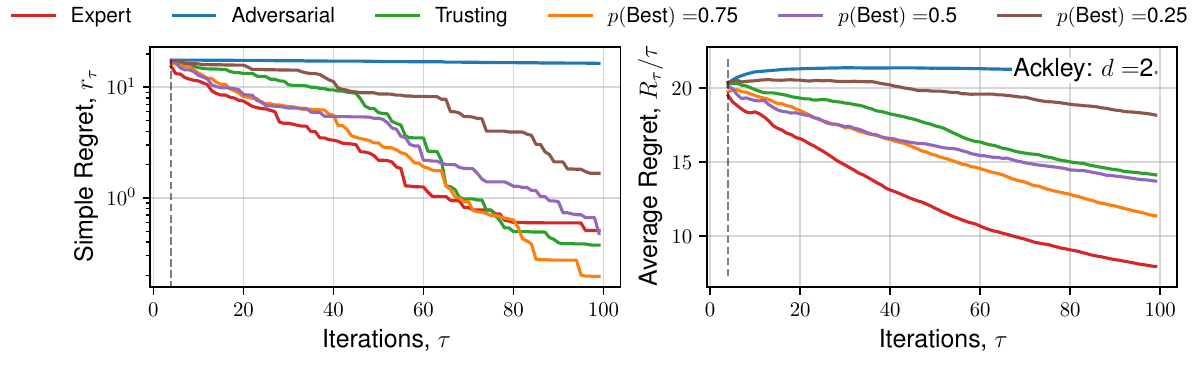}
    \caption{2D Ackley function.}
    \label{example_ackley2}
\end{figure}

\begin{figure}[htb!]
    \centering
    \includegraphics[width=\textwidth]{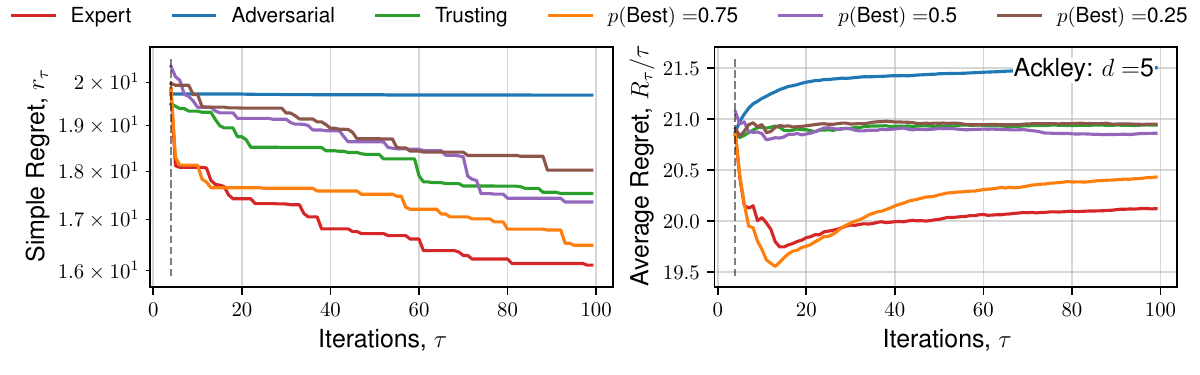}
    \caption{5D Ackley function.}
    \label{example_ackley5}
\end{figure}

\begin{figure}[htb!]
    \centering
    \includegraphics[width=\textwidth]{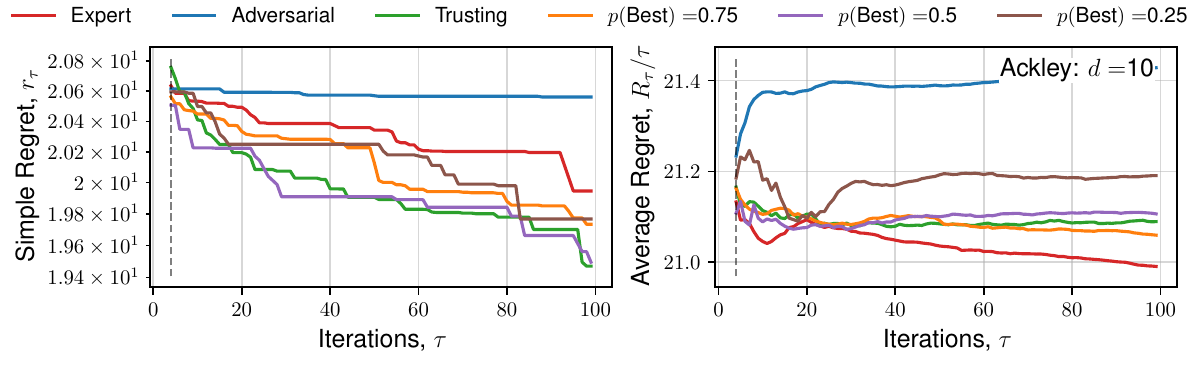}
    \caption{10D Ackley function.}
    \label{example_ackley10}
\end{figure}

\begin{figure}[htb!]
    \centering
    \includegraphics[width=\textwidth]{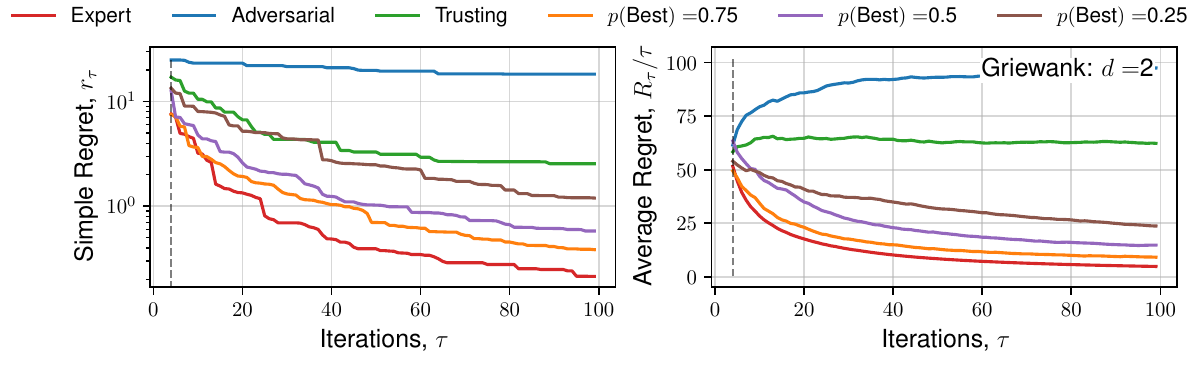}
    \caption{2D Griewank function.}
    \label{example_Griewank2}
\end{figure}

\begin{figure}[htb!]
    \centering
    \includegraphics[width=\textwidth]{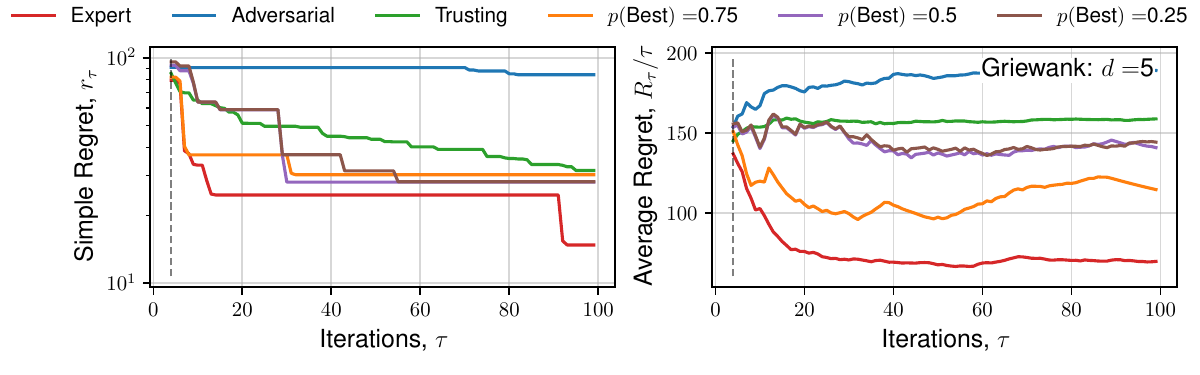}
    \caption{5D Griewank function.}
    \label{example_Griewank5}
\end{figure}

\begin{figure}[htb!]
    \centering
    \includegraphics[width=\textwidth]{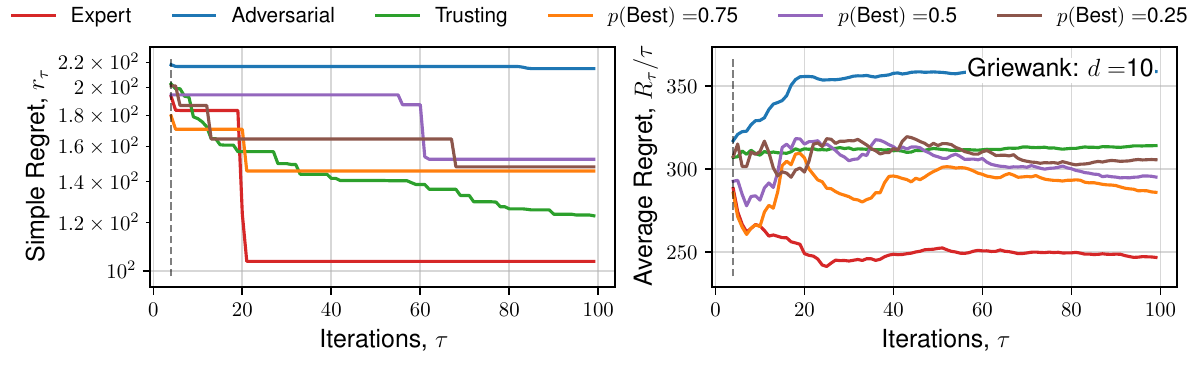}
    \caption{10D Griewank function.}
    \label{example_Griewank10}
\end{figure}

\begin{figure}[htb!]
    \centering
    \includegraphics[width=\textwidth]{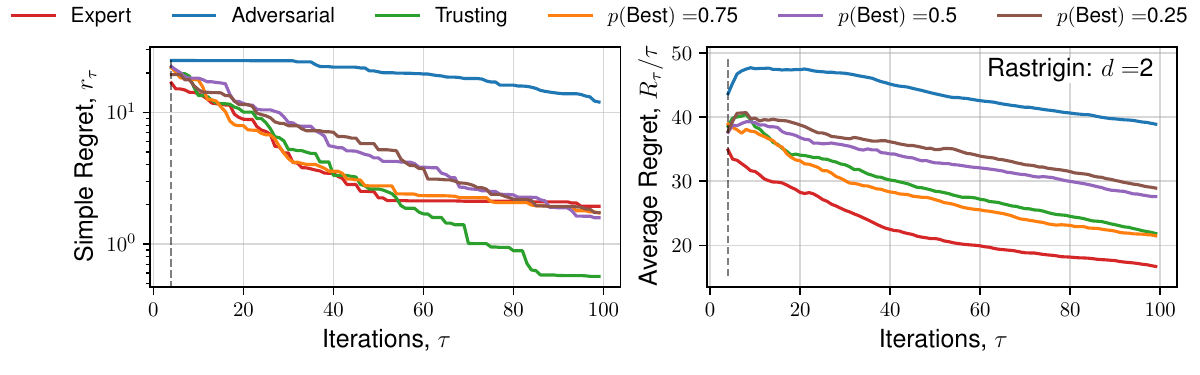}
    \caption{2D Rastrigin function.}
    \label{example_Rastrigin2}
\end{figure}

\begin{figure}[htb!]
    \centering
    \includegraphics[width=\textwidth]{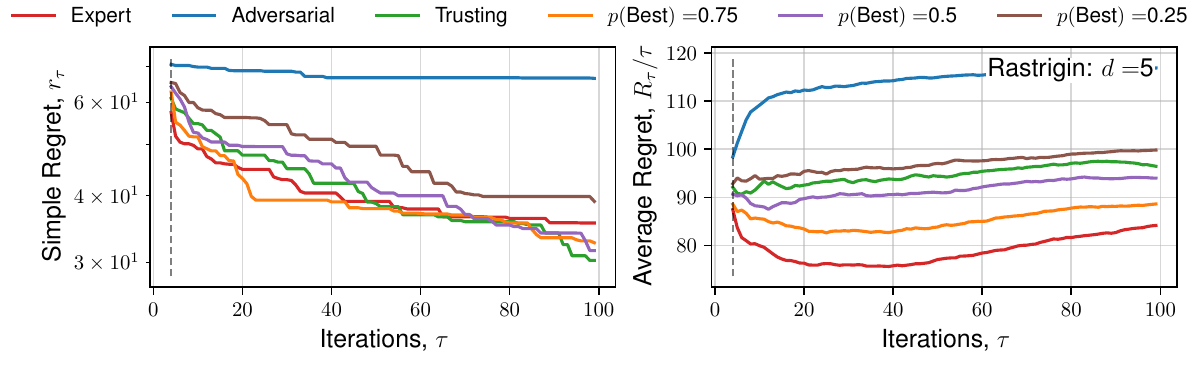}
    \caption{5D Rastrigin function.}
    \label{example_Rastrigin5}
\end{figure}

\begin{figure}[htb!]
    \centering
    \includegraphics[width=\textwidth]{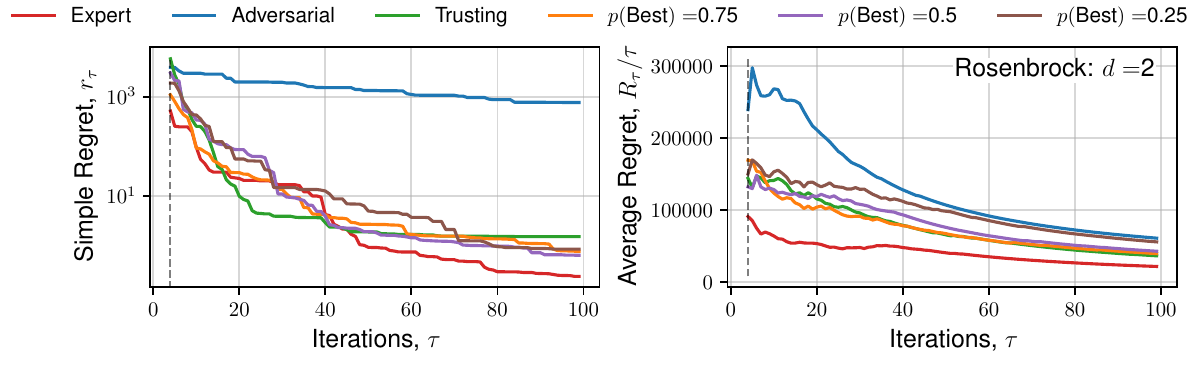}
    \caption{2D Rosenbrock function.}
    \label{example_Rosenbrock2}
\end{figure}

\begin{figure}[htb!]
    \centering
    \includegraphics[width=\textwidth]{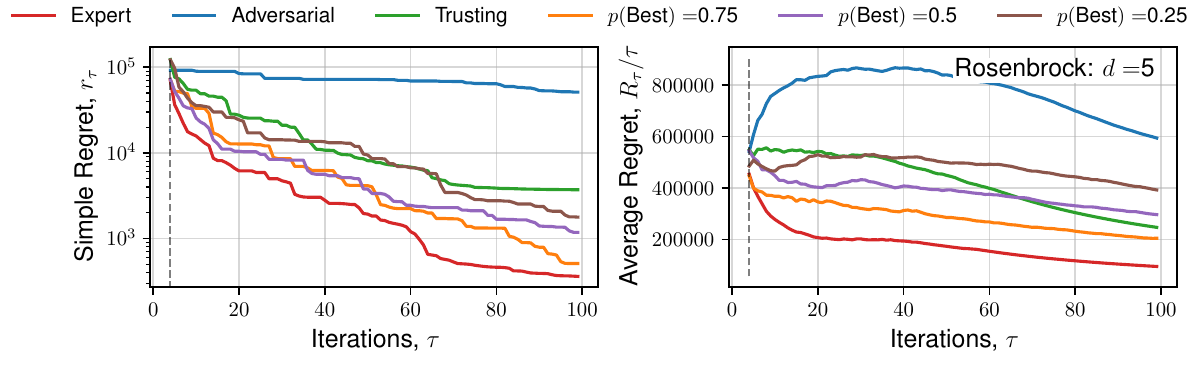}
    \caption{5D Rosenbrock function.}
    \label{example_Rosenbrock5}
\end{figure}

\begin{figure}[htb!]
    \centering
    \includegraphics[width=\textwidth]{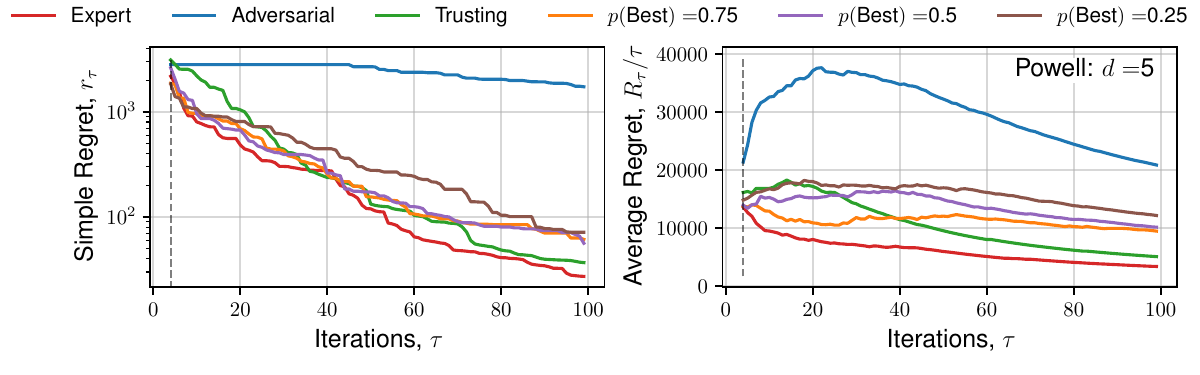}
    \caption{5D Powell function.}
    \label{example_Powell5}
\end{figure}

\end{document}